# Parsing of part-of-speech tagged Assamese Texts

**Mirzanur Rahman[1], Sufal Das[1] and Utpal Sharma[2]**

[1] Department of Information Technology, Sikkim Manipal Institute of Technology
Rangpo, Sikkim-737136, India

[2] Department of Computer Science & Engineering, Tezpur University
Tezpur, Assam-784028, India

**Abstract**

A natural language (or ordinary language) is a language that is spoken, written, or signed by humans for general-purpose communication, as distinguished from formal languages (such as computer-programming languages or the "languages" used in the study of formal logic). The computational activities required for enabling a computer to carry out information processing using natural language is called natural language processing. We have taken Assamese language to check the grammars of the input sentence. Our aim is to produce a technique to check the grammatical structures of the sentences in Assamese text. We have made grammar rules by analyzing the structures of Assamese sentences. Our parsing program finds the grammatical errors, if any, in the Assamese sentence. If there is no error, the program will generate the parse tree for the Assamese sentence

## 1. Introduction

Natural language processing, a branch of artificial intelligence that deals with analyzing, understanding and generating the languages that humans use naturally in order to interface with computers in both written and spoken contexts using natural human languages instead of computer languages. It studies the problems of automated generation and understanding of natural human languages. We have taken Assamese language for information processing i.e. to check the grammars of the input sentence. Parsing process makes use of two components. A parser, which is a procedural component and a grammar, which is declarative. The grammar changes depending on the language to be parsed while the parser remains unchanged. Thus by simply changing the grammar, the system would parsed a different language. We have taken Earley's Parsing Algorithm for parsing Assamese Sentence according to a grammar which is defined for Assamese language.

## 2. Related Works

### 2.1 Natural Language Processing

The term "natural" languages refer to the languages that people speak, like English, Assamese and Hindi etc. The goal of the Natural Language Processing (NLP) group is to design and build software that will analyze, understand, and generate languages that humans use naturally. The applications of Natural Language can be divided into two classes [2]

- Text based Applications: It involves the processing of written text, such as books, newspapers, reports, manuals, e-mail messages etc. These are all reading based tasks.
- Dialogue based Applications: It involves human – machine communication like spoken language. Also includes interaction using keyboards. From an end-user's perspective, an application may require NLP for either processing natural language input or producing natural language output, or both. Also, for a particular application, only some of the tasks of NLP may be required, and depth of analysis at the various levels may vary. Achieving human like language processing capability is a difficult goal for a machine.

The difficulties are:
- Ambiguity
- Interpreting partial information
- Many inputs can mean same thing

### 2.2 Knowledge Required for Natural Language

A Natural Language system uses the knowledge about the structure of the language itself, which includes words and how words combine to form sentences, about the word meaning and how word meanings contribute to sentence meanings and so on. The different forms of knowledge relevant for natural language are [2]:

- Phonetic and phonological knowledge: It concerns how words are related to the sounds that realize them.
- Morphological knowledge: It concerns how words are constructed from more basic meaning units called morphemes. A morpheme is the primitive unit of meaning in a language (for example, the meaning of the word friendly is derivable from the meaning of the noun friend and suffix -ly, which transforms a noun into an adjective.
- Syntactic Knowledge: It concerns how words can be put together to form correct sentences and determine what structure role each word plays in the sentence.
- Semantic knowledge: It concerns what words mean and how these meanings combine in sentences to form sentence meanings.
- Pragmatic Knowledge: It concerns how sentences are used in different situations and how use affects the interpretation of the sentence.
- Discourse Knowledge: It concerns how the immediately preceding sentences affect the interpretation of the next sentence.
- Word Knowledge: It includes what each language user must know about the other user's beliefs and goals.

2.3 Earley's Parsing Algorithm

The Earley's Parsing Algorithm [8, 9] is basically a top down parsing algorithm where all the possible parses are carried simultaneously. Earley's algorithm uses dotted Context-free Grammar (CFG) rules called items, which has a dot in its right hand side.
Let the input sentence be- "0 I 1 saw 2 a 3 man 4 in 5 the 6 park 7". Here the numbers appeared between words are called position numbers.
For CFG rule S →NP VP we will have three types of dotted items-
• [ S→ .NP VP,0,0 ]
• [ S→ NP.VP,0,1 ]
• [ S→ NP VP.,0,4 ]

Here
$S \rightarrow$ Starting Symbol
$NP \rightarrow$ Noun Phrase
$VP \rightarrow$ Verb Phrase

1. The first item indicates that the input sentence is going to be parsed applying the rule S→ NP VP from position 0.
2. The second item indicates the portion of the input sentence from the position number 0 to 1 has been parsed as NP and the remainder left to be satisfied as VP.
3. The third item indicates that the portion of input sentence from position number 0 to 4 has been parsed as NP VP and thus S is accomplished.

Earley's algorithm uses 3 phases
- Predictor
- Scanner
- Completer

Let α, β, γ are sequence of terminal or nonterminal symbols and S, A, B are non terminal symbols.

**Predictor Operation**
For an item of the form [A → α.Bβ,i,j] create [B→.γ,j,j] for each production of the [B→γ] It is called predictor operation because we can predict the next item.

**Completer Operation**
For an item of the form [B→γ.,j,k] create [A→αB.β,i,k] (i<j<k) for each item in the form of [A→α.Bβ,i,j] if exists. It is called completer because it completes an operation.

**Scanner Operation**
For an item of the form [A→α.wβ,i,j] create [A→αw.β,i,j+1], if w is a terminal symbol appeared in the input sentence between j and j+1.

**Earley's parsing algorithm**

1. For each production S→α, create [S→.α,0,0]
2. For j=0 to n do (n is the length of the input sentence)
3. For each item in the form of [A→α.Bβ,i,j] apply Predictor operation while a new item is created.
4. For each item in the form of [B→γ.i,j] apply Completer operation while a new item is created.
5. For each item in the form of [A→α.wβ,i,j] apply Scanner operation

If we find an item of the form [S→α.,0,n] then we accept it.

Let us take an example..
"0 I 1 saw 2 a 3 man 4".
Consider the following grammar:
1. S → NP VP
2. S → S PP
3. NP → n

4. NP → art n
5. NP → NP PP
6. PP → p NP
7. VP → v NP
8. n → I
9. n → man
11. v → saw
12. art → a

Now parse the sentence using Earley's parsing technique.

| 1. | [S→.NP VP,0,0] | Initialization |
| 2. | [S→.S PP,0,0] | Apply Predictor to step 1 and step 2 |
| 3. | [NP→.n,0,0] | |
| 4. | [NP→.art n,0,0] | |
| 5. | [NP→.NP PP,0,0] | Apply Predictor to step 3 |
| 6. | [n→."I",0,0] | Apply scanner to 6 |
| 7. | [n→ "I".,0,1] | Apply Completer to step 7 with step 3 |
| 8. | [NP→n.,0,1] | Apply Completer to step 8 with step 1 and step 5 |
| 9. | [S→NP.VP,0,1] | |
| 10. | [NP→NP.PP,0,1] | Apply Predictor to step 9 |
| 11. | [VP→.v NP,1,1] | Apply Predictor to step 11 |
| 12. | [v→."saw",1,1] | Apply Predictor to step 10 |
| 13. | [PP→.p NP,1,1] | Apply Scanner to step 12 |
| 14. | [v→ "saw".1,2] | Apply Completer to step 14 with step 11 |
| 15. | [VP→v.NP,1,2] | Apply Predictor to step 15 |
| 16. | [NP→.n,2,2] | |
| 17. | [NP→.art n,2,2] | |
| 18. | [NP→.NP PP 2,2] | Apply Predictor to step 17 |
| 19. | [art → ."a", 2,2] | Apply Scanner to step 19 |
| 20. | [art → "a".,2,3] | Apply Completer to step 20 with step 17 |
| 21. | [NP → art .n,2,3] | Apply Predictor to step 21 |
| 22. | [n → ."man", 3,3] | Apply Scanner to step 22 |
| 23. | [n → "man".,3,4] | Apply Completer to step 23 with step 21 |
| 24. | [NP → art n.,2,4] | Apply Completer to 24 with 15 |
| 25. | [VP → v NP.,1,4] | Apply Completer to 25 with 9 |
| 26. | [S → NP VP.,0,4] | Complete |

. When applying Predictor operation Earley's algorithm often creates a set of similar items such as-step 3,4,5 and 16,17,18 expecting NP in future.

## 3. Properties and problems of parsing algorithm

Parsing algorithms are usually designed for classes of grammar rather than for some individual grammars. There are some important properties [6] that make a parsing algorithm practically useful.

- It should be sound with respect to a given grammar and lexicon
- It should be complete so that it assign to an input sentence and all the analyses it can have with respect to the current grammar and lexicon.
- It should also be efficient so that it take minimum of computational work.

Algorithm should be robust, behaving in a reasonably sensible way when presented with sentence that it is unable to fully analyze successfully.

The main problem of Natural Language is its ambiguity. The sentences of Natural Languages are ambiguous in meaning. There are different meanings for one sentence. So all the algorithms for parsing can not be used for Natural Language processing. There are many parsing technique used in programming languages (like C language).These techniques easy to use, because in programming language, meaning of the words are fixed. But in case of NLP we can not used this technique for parsing, because of ambiguity.

For example: "I saw a man in the park with a telescope".

This sentence has at least three meanings-
- Using a telescope I saw the man in the park.
- I saw the man in the park that has a telescope.
- I saw the man in the park standing behind the telescope which is placed in the park.

So, this sentence is ambiguous and no algorithm can resolve the ambiguity. An algorithm will be the best algorithm, which produces all the possible analyses.
To begin with, we look for algorithm that can take care of ambiguity of smaller components such as ambiguity of words and phrases.

## 4. Proposed grammar and algorithm for Assamese Texts

Since it is impossible to cover all types of sentences in Assamese language, we have taken some portion of the sentence and try to make grammar for them. Assamese is free-word-order language [10]. As an example we can take the following Assamese sentence.

মই পার্কত মানুহ এজন দেখিছো
("mai pArkt mAnuH ezan dekhiCo")

This sentence can be written as.

মই মানুহ এজন পাৰ্কত দেখিছো    (PN - NP - ART - NP - VP)
মই এজন মানুহ পাৰ্কত দেখিছো    (PN - ART - NP - NP - VP)
মানুহ এজন মই পাৰ্কত দেখিছো    (NP - ART - PN - NP -VP)
পাৰ্কত মানুহ এজন মই দেখিছো    (NP - NP - ART - PN - NP)
এজন মানুহ মই পাৰ্কত দেখিছো    (ART- NP - PN - NP - VP)
পাৰ্কত মই মানুহ এজন দেখিছো    (NP - PN - NP - ART - VP)

Here we see that one sentence can be written in different forms for the same meaning, i.e. the positions of the tags are not fixed. So we can not restrict the grammar rule for one sentence. The grammar rule may be very long, but we have to accept it. The grammar rule we have tried to make, may not work for all the sentences in Assamese language. Because we have not considered all types of sentences. Some of the sentences are shown below, which are used to make the grammar rule [3, 4].

1. নৈ খন বৰ ডাঙৰ    (NP-ART-ADJ-ADJ)
2. ঘোঁৰাটোৱে ঘাহ খাইছে    (NP-NP-VP)
3. দিল্লী এখন মহানগৰ    (NP-ART-NP)
4. আজি বৰষুণ দিছে    (PN-NP-VP)
5. মই আৰু সি একেলগে ঘৰলৈ যাম    (PN-IND-PN-ADV-NP-VP)
6. বাঃ কি সুন্দৰ ফুল    (IND-ADV-ADJ-NP)
7. হেৰা এইফালে আহাছোন    (IND-PN-VP)
8. তেওঁ হলে কৰিলেহেতেন    (PN-IND-VP)
9. শিক্ষকে ছাত্ৰক কিতাপখন দিলে    (NP-NP-NP-VP)
10. খৰকৈ যোৱা    (ADV-VP)
11. অতি গুণৱতী ছোৱালী    (ADJ-ADJ-NP)
12. পকা কঠাল    (ADJ-NP)
13. সোনকালে আহিবা    (ADJ-VP)
14. গৰু এবিধ উপকাৰী জন্তু    (NP-ART-ADJ-NP)

**Our proposed grammars for Assamese sentences**

1. S → PP VP | PP
2. PP → PN NP | NP PN | ADJ NP | NP ADJ | NP | ADJ | IND NP | PN | ADV NP | ADV
3. NP → NP PP | PP NP | ADV NP | PP | ART NP | NP ART | IND PN | PN IND

Here.....
NP → Noun
PN → Pronoun
VP → Verb
ADV → Adverb
ADJ → Adjective
ART → Article
IND → Indeclinable

**4.1 Modification of Earley's Algorithm for Assamese Text Parsing**

We know that Earley's algorithm uses three operations, Predictor, Scanner and Completer. We add Predictor and Completer in one phase and Scanner operation in another phase.
Let α, β, γ, PP, VP are sequence of terminal or nonterminal symbols and S, B are non terminal symbols.

**Phase 1:(Predictor+Completer)**
For an item of the form [S →α .Bβ,i,j] , create
[S →α.γβ,i,j] for each production of the [B→γ]

**Phase 2 :( Scanner)**
For an item of the form [S→α.wβ,i,j] create [S→αw.β,i,j+1], if w is a terminal symbol appeared in the input sentence between j and j+1.

**Our Algorithm**
Input: Tagged Assamese Sentence
Output: Parse Tree or Error message
Step 1: If Verb is present in the sentence then create
       [S→ .PP VP ,0,0]
       Else create
       [S→ .PP ,0,0]
Step 2: Do the following steps in a loop until there is a success or error
Step 3: For each item of the form of [S→α.Bβ,i,j], apply phase 1
Step 4: For each item of the form of [S→ .αwβ,i,j], apply phase 2
Step 5: If we find an item of the form [S→α. ,0,n], then we accept the sentence as success else error message. Where n is the length of input sentence. And then come out from the loop.
Step 6: Generate the parse trees for the successful sentences.

**Some other modifications of Earley's algorithm:**

1. Earley's algorithm blocks left recursive rules [NP→ .NP PP ,0,0], when applying Predictor operation. Since Assamese Language is a Free-Word-Order language. We are not blocking this type of rules.
2. Earley's algorithm creates new items for all possible productions, if there is a non terminal in the left hand side rule. But we reduce these productions by removing such type of productions, which create the number of total productions in the stack, greater then total tag length of the input sentence.
3. Another restriction we used in our algorithm for creating new item is that, if the algorithm currently analyzing the last word of the sentence, then it selects only the single production in the right hand side (example [PP→NP]). The other rules (which have more then one production rules in right hand side (example [PP→PN NP])) are ignored by the algorithm.

4.2 Parsing Assamese text using proposed grammar and algorithm

Let us take an Assamese sentence.

মই আৰু সি একেলগে ঘৰলৈ যাম

( " mai Aru si ekelge gharalE jAm" )

Now the position number for the words are placed according to which word will be parsed first.

0 mai 1 Aru 2 si 3 ekelge 4 gharalE 5 jAm 6
0 মই 1 আৰু 2 সি 3 একেলগে 4 ঘৰলৈ 5 যাম 6

We consider the following grammar rule
1. S → PP VP | PP
2. PP → PN NP | NP PN | ADJ NP | NP ADJ | NP | ADJ | IND NP | PN | ADV NP | ADV
3. NP → NP PP | PP NP | ADV NP | PP | ART NP | NP ART | IND PN | PN IND
4. PN → "mai"
5. PN → "si"
6. IND → "Aru"
7. ADV → "ekelge"
8. NP → "gharalE"
9. VP → "jAm"

Parsing process will proceed as follows

| 1 | [S → .PP VP , 0,0] | Apply Phase 1 |
| 2 | [S → .NP VP,0 ,0] | Apply Phase 1 |
| 3 | [S → .PP NP VP,0,0] | Apply Phase 1 |
| 4 | [S → .PN NP NP VP,0,0] | Apply Phase 1 |
| 5 | [S → ."mai" NP NP VP,0,0] | Apply Phase 2 |
| 6 | [S → ."mai" .NP NP VP,0,1] | Apply Phase 1 |
| 7 | [S → "mai" .IND PN NP VP,0,1] | Apply Phase 1 |
| 8 | [S → "mai" . "Aru" PN NP VP,0,1] | Apply Phase 2 |
| 9 | [S → "mai" "Aru" .PN NP VP,0,2] | Apply Phase 1 |
| 10 | [S → "mai" "Aru" ."si" NP VP,0,2] | Apply Phase 2 |
| 11 | [S → "mai" "Aru" "si" .NP VP,0,3] | Apply Phase 1 |
| 12 | [S → "mai" "Aru" "si" .ADV NP VP,0,3] | Apply Phase 1 |
| 13 | [S → "mai" "Aru" "si" ."ekelge" NP VP,0,3] | Apply Phase 2 |
| 14 | [S → "mai" "Aru" "si" "ekelge" .NP VP,0,4] | Apply Phase 1 |
| 15 | [S → "mai" "Aru" "si" "ekelge" ."gharalE" .VP,0,5] | Apply Phase 2 |
| 16 | [S → "mai" "Aru" "si" "ekelge" "gharalE" .VP,0,5] | Apply Phase 1 |
| 17 | [S → "mai" "Aru" "si" "ekelge" "gharalE" ."jAm",0,5] | Apply Phase 2 |
| 18 | [S → "mai" "Aru" "si" "ekelge" "gharalE" "jAm".,0,6] | Complete |

In the above example, we have shown only the steps which proceeds to the goal. The other steps are ignored.

## 5. Implementation and Result Analysis

5.1 Different Stages of the Program

In the program there are 3 stages.
- Lexical Analysis
- Syntax Analysis
- Tree Generation

In Lexical Analysis stage, program finds the correct tag for each word in the sentence by searching the database. There are seven databases (NP, PN, VP, ADJ, ADV, ART, IND) for tagging the words.

In Syntax Analysis stage the program tries to analyze whether the given sentence is grammatically correct or not.

In Tree Generation stage, the program finds all the production rules which lead to success and generates parse tree for those rules. If there are more then one path to success, this stage can generates more then on parse trees. It also displays the words of the sentences with proper tags. The following shows a parse tree generate by the program.

```
INPUT SENTENCE--> : mai Aru si ekelge gharalE jAm.

SENTENCE RECOGNIZED

TREE--->
1.  S
2.  [S --> (PP VP )]
3.  [PP --> (NP )]VP
4.  [NP --> (PP NP )]VP
5.  [PP --> (PN NP )]NP VP
6.  [PN --> (pn  : mai)]NP NP VP
7.  [NP ]NP VP
8.  [NP --> (IND PN )]NP VP
9.  [IND --> (ind  : Aru)]PN NP VP
10. [PN ]NP VP
11. [PN --> (pn  : si)]NP VP
12. [NP ]VP
13. [NP --> (ADV NP )]VP
14. [ADV --> (adv  : ekelge)]NP VP
15. [NP ]VP
16. [NP --> (np  : gharalE)]VP
17. [VP ]
18. [VP --> (vp  : jAm)]
```

The original parse tree for the above sentence is.

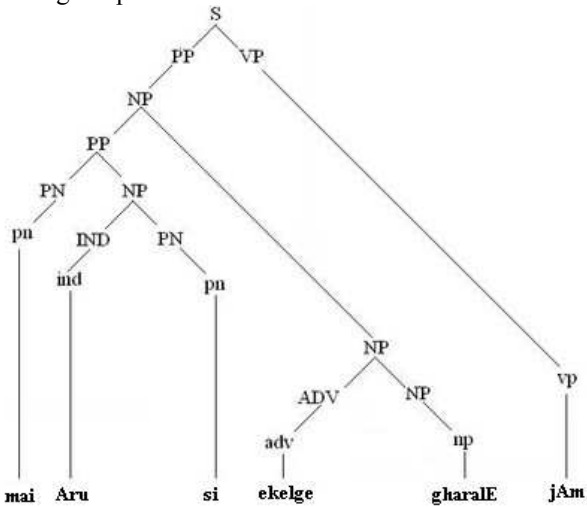

### 5.2 Result Analysis

After implementation of Earley's algorithm using our proposed grammar, it has been seen that the algorithm can easily generates parse tree for a sentence if the sentence structure satisfies the grammar rules. For example we take the following Assamese sentence

গৰু এবিধ উপকাৰী জন্তু (gru ebidh upakArI za\ntu).

The structure of the above sentence is NP-ART-ADJ-NP. This is a correct sentence according to the Assamese literature. According to our proposed grammar a possible top down derivation for the above sentence is

| 1.  | S                          | [Handle]       |
|-----|----------------------------|----------------|
| 2.  | >>PP                       | [S→PP]         |
| 3.  | >> NP                      | [PP→NP]        |
| 4.  | >>NP PP                    | [NP→NP PP]     |
| 5.  | >>NP ART PP                | [NP → NP ART]  |
| 6.  | >>gru ART PP               | [NP → gru      |
| 7.  | >>gru ebidh PP             | [ART→ ebidh    |
| 8.  | >>gru ebidh  ADJ NP        | [PP→ ADJ NP]   |
| 9.  | >>gru ebidh  upakArI NP    | [ADJ→upakArI]  |
| 10. | >>gru ebidh  upakArI za\ntu| [NP→za\ntu]    |

From the above derivation it has been seen that the Assamese sentence is correct according to the proposed grammar. So our parsing program generates a parse tree successfully as follows.

```
INPUT SENTENCE--> : gru ebidh upakArI za\ntu.
SENTENCE RECOGNIZED
TREE--->
1.  S
2.  [S --> (PP )]
3.  [PP --> (NP )]
4.  [NP --> (NP PP )]
5.  [NP --> (NP ART )]PP
6.  [NP --> (np  : gru)]ART PP
7.  [ART ]PP
8.  [ART --> (art  : ebidh)]PP
9.  [PP ]
10. [PP --> (ADJ NP )]
11. [ADJ --> (adj  : upakArI)]NP
12. [NP ]
13. [NP --> (np  : za\ntu)]
```

Our program tests only the sentence structure according to the proposed grammar rules. So if the sentence structure satisfies the grammar rule, program recognizes the sentence as a correct sentence and generates parse tree. Otherwise it gives output as an error.

## 6. Conclusion and Future Work

We have developed a context free grammar for simple Assamese sentences. Different natural languages present different challenges in computational processing. We have studied the issues that arise in parsing Assamese sentences and produce an algorithm suitable for those issues. This algorithm is a modification of Earley's Algorithm. We found that Earley's parsing algorithms is simple and effective.

In this work we have considered limited number of Assamese sentences to construct the grammar rules. We also have considered only seven main tags. In future work we have to consider as many sentences as we can and some more tags for constructing the grammar rules. Because Assamese language is a free-word-order language. Word position for one sentence may not be same in the other sentences. So we can not restrict the grammar rules for some limited number of sentences.